\pdfoutput=1
\documentclass[10pt,twocolumn,letterpaper]{article}

\usepackage[pagenumbers]{cvpr} 

\newcommand{\methodName}{H-MoRe}
\newtoggle{draft}
\togglefalse{draft}

\usepackage{ulem}
\definecolor{carmine}{rgb}{0.59, 0.0, 0.09}
\definecolor{turquoise}{rgb}{0.25, 0.88, 0.82}
\definecolor{teal}{rgb}{0.0, 0.5, 0.5}
\definecolor{cinereous}{rgb}{0.6, 0.51, 0.48}
\definecolor{darkblue}{rgb}{0.129, 0.255, 0.451}
\definecolor{darkgreen}{rgb}{0.165, 0.373, 0.357}
\definecolor{darkpink}{rgb}{0.494, 0.122, 0.278}
\definecolor{mint}{rgb}{0.92, 0.96, 0.90}
\definecolor{pink}{rgb}{0.85, 0.60, 0.70}
\definecolor{orange}{rgb}{0.85, 0.44, 0.05}

\iftoggle{draft} {
    \newcommand{\com}[1]{\textcolor{red}{[#1]}}
    \newcommand{\mfinish}[1]{\textbf{\color{teal}[#1]}}
    \newcommand{\mstatus}[1]{\textbf{\color{carmine}[#1]}}
} {
    \newcommand{\com}[1]{}
    \newcommand{\mfinish}[1]{}
    \newcommand{\mstatus}[1]{}
}

\usepackage{tikz}
\usepackage{afterpage}
\usepackage{placeins}
\usepackage{mathtools}
\usepackage{array}
\usepackage{multirow}
\usepackage{makecell}
\usepackage{tabularx}

\newcommand{\msubcaption}[1]{#1} 
\newcommand{\mcaption}[1]{\textbf{#1}}
\newcommand{\mvspace}{} 

\definecolor{cvprblue}{rgb}{0.21,0.49,0.74}
\usepackage[pagebackref,breaklinks,colorlinks,allcolors=cvprblue]{hyperref}

\def\paperID{1583} 
\def\confName{CVPR}
\def\confYear{2025}

\title{H-MoRe: Learning Human-centric Motion Representation for Action Analysis}

\author{Zhanbo Huang, Xiaoming Liu, Yu Kong \\
{\normalsize Department of Computer Science and Engineering, Michigan State University} \\
{\tt\small \{huang247\}@msu.edu, \{liuxm,yukong\}@cse.msu.edu}
}

\begin{document}
\maketitle

\begin{abstract}
In this paper, we propose \methodName{}, a novel pipeline for learning precise human-centric motion representation. 
Our approach dynamically preserves relevant human motion while filtering out background movement. 
Notably, unlike previous methods relying on fully supervised learning from synthetic data, \methodName{} learns directly from real-world scenarios in a self-supervised manner, incorporating both human pose and body shape information.
Inspired by kinematics, \methodName{} represents absolute and relative movements of each body point in a matrix format that captures nuanced motion details, termed world-local flows.
\methodName{} offers refined insights into human motion, which can be integrated seamlessly into various action-related applications.
Experimental results demonstrate that \methodName{} brings substantial improvements across various downstream tasks, including gait recognition~{\small(CL@R1: 16.01\%\(\uparrow\))}, action recognition~{\small(Acc@1: 8.92\%\(\uparrow\))}, and video generation~{\small(FVD: 67.07\%\(\downarrow\))}. 
Additionally, \methodName{} exhibits high inference efficiency (\(34\) fps), making it suitable for most real-time scenarios.
Models and code is available at \url{https://github.com/haku-huang/h-more}.
\end{abstract}
\section{Introduction {\small \mfinish{Nov.9} \mstatus{Ready}}} \label{sec:introduction}

Understanding human motion has long been a fundamental challenge in computer vision.
Leveraging motion information enables numerous practical applications, from rehabilitation~\cite{gu2019home, francisco2022computer} in healthcare to video surveillance~\cite{pu2024learning, zhu2017flow} in security. This analysis relies on recognizing patterns in human motion, rather than the movement of irrelevant background elements (\eg, tree leaves).
To enhance analysis performance, it is essential to extract precise human-centric motion that captures the nuances of human movement while suppressing irrelevant background dynamics.

\begin{figure}[t]
    \centering
    \includegraphics[width=0.48\textwidth]{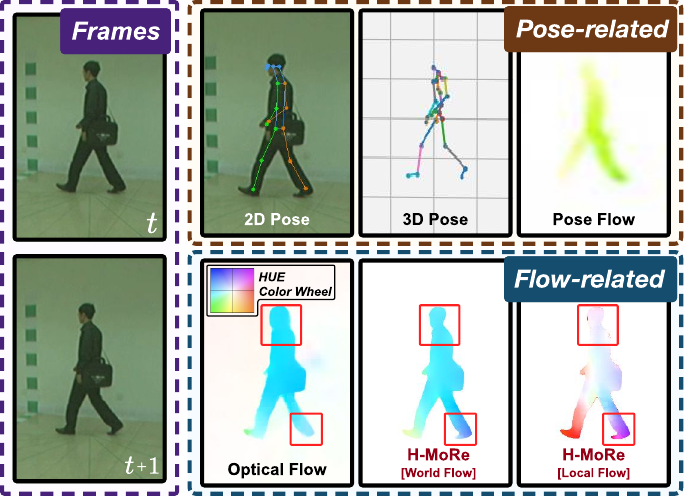}
    \caption{\mcaption{Comparison of our \methodName{} with other motion representations.} \msubcaption{We visualize three pose-related representations -- 2D Pose~\cite{fang2022alphapose}, 3D Pose~\cite{sarandi2023dozens}, and PoseFlow~\cite{zhang2018poseflow} -- as well as flow-related representations: Optical Flow~\cite{teed2020raft}, and our \methodName{} (contains world flow and local flow). The  red box highlights \methodName{}'s precise motion information and sharp boundaries.}}
    \label{fig:first_figure}
\end{figure}

Currently, human motion is commonly represented using two methods: optical flow and human pose.
Optical flow captures motion between consecutive frames using a flow map to represent pixel motion offsets.
It encodes both motion and shape information~\cite{baker2011database, weinzaepfel2015learning} in a matrix format, making it easily perceptible by Convolutional Neural Networks (CNNs) or Vision Transformers (ViTs).
As a result, optical flow has become a popular input signal for action recognition~\cite{pu2022semantic, sun2018optical, bao2021evidential} and video generation~\cite{yang2023rerender, hu2023dynamic, xing2023dynamicrafter}.
However, it calculates offsets indiscriminately for all pixels, which is inadequate for human-centric scenarios with moving backgrounds.
Additionally, optical flow learned from synthetic datasets lacks real biological entities~\cite{ilg2017flownet, mayer2016large, vogel20153d} and struggles to capture subtle nuances of human motion.
Consequently,  recent analysis tasks~\cite{zaier2023dual, liao2020model} have shifted to using human pose, which represents motion through 2D~\cite{jiang2023rtmpose, yang2023effective} or 3D skeletal points~\cite{zheng20213d, zhou2019hemlets, baradel2022posebert}. 
However, human pose lacks body shape information, which is particularly important for specific tasks such as gait recognition~\cite{zhu2021gait, cheng2017discriminative}.

To address these challenges, we propose \textbf{\methodName{}}, a new formulation for \textbf{H}uman-centric \textbf{Mo}tion \textbf{Re}presentation that accurately preserves both the motion and shape information of the human body.
Unlike the Endpoint Error (EPE)~\cite{fleet1990computation}, a common objective function for fully-supervised optical flow estimation, we learn \methodName{} directly from real-world scenarios through a \emph{joint constraint learning framework}, which incorporates two self-supervised constraints -- a skeleton constraint and a boundary constraint -- to capture accurate motion and shape information, respectively.
The skeleton constraint establishes associations between each body point and the skeleton, allowing \methodName{} to use skeleton offsets to approximate the range of body motion. 
Meanwhile, the boundary constraint refines human motion in local regions, such as hands and feet, and ensures that \methodName{} captures the overall body shape.
This is achieved by aligning the edges in our representation with human boundaries during training.
The \emph{joint constraint learning framework} enables \methodName{} to enhance performance across various human-centric tasks that heavily rely on motion and shape information, such as gait recognition~\cite{fan2020gaitpart, lin2021gait, ye2024biggait,zhang2019gait}, action recognition~\cite{dhiman2020view, lin2023video, zhao2023search}, and video generation~\cite{guo2022action2video, gomes2021shape, zhang2024mimicmotion}.

Additionally, while most motion representations depict human movement within an environmental context, certain tasks~\cite{zheng2023parsing, girase2023latency, zhao2023antgpt} require focused analysis of specific body parts, \eg, the left arm.
Drawing on principles of physical kinematics, \methodName{} introduces local flow to represent movements relative to the subject, highlighting intrinsic motion details.
Combined with world flow, which captures the movement of each body point relative to the environment, this \emph{world-local flows} provides a more comprehensive and nuanced representation of human motion, incorporating both absolute and relative movement information for downstream tasks.

\cref{fig:first_figure} highlights the unique features of our world-local flows. 
By visualizing flow using the HUE color wheel~\cite{barron1994performance}, we can clearly see that world flow captures motion details with exceptional clarity, including body boundaries and finer points like foot movement -- outperforming existing methods.
Meanwhile, local flow reveals the distinctive qualities of local flow, with a richer color spectrum indicating more detailed underlying motion information.
This self-referential flow provides unique insights for downstream tasks involving human motion.

Our contributions are summarized as follows:
\begin{enumerate}
    \item We introduce an innovative pipeline, termed \methodName{}, which provides a human-centric motion representation with precise motion and shape information, easily integrable into any motion analysis algorithms.
    \item We design the \emph{joint constraint learning framework}, enabling the estimation model to learn human motion from any real-world sequences without ground truth.
    \item We propose \emph{world-local flows}, which capture absolute and relative motions and provide richer insights about human-centric motion for practical analysis tasks.
    \item Extensive experiments on three distinct analysis tasks demonstrate the superiority of our \methodName{} in enhancing the performance of various action analysis tasks.
\end{enumerate}

\section{Related Work {\small \mfinish{Oct.9} \mstatus{Ready}}} \label{sec:related_work}


\subsection{Learning Methods of Optical Flow}
Initially, optical flow estimation is formulated as an energy minimization problem~\cite{horn1981determining}. 
This problem is constructed using such combined regularization terms by assuming that moving objects maintain the same brightness across frames and the flow is smooth between adjacent pixels.
Subsequently, with the development of CNNs, the Endpoint Error (EPE) is introduced to calculate the Euclidean distance between predicted and ground truth at the pixel level. 
Many optical flow estimation methods use EPE as the supervisory signal at single~\cite{dosovitskiy2015flownet} or multiple scales~\cite{teed2020raft, sui2022craft}.
Recently, thanks to the global information extraction capability of Transformers, some methods~\cite{xu2022gmflow, huang2022flowformer} construct loss functions based on the matching cost between each pixel pair in the feature space, termed Cost Volume Loss. 
It is calculated by measuring the deviation between the minimum cost location in the cost volume and ground truth.
Further, FlowFormer{\scriptsize++}~\cite{shi2023flowformer++} enhances the model's understanding of global features by applying a masking mechanism to the cost volume.
However, these pixel-level loss functions are often ineffective in estimating human-centric motion. 
Thus, exploring a human-centric learning framework to learn human motion from real-world scenarios is particularly desirable.
To address this, we propose the \emph{joint constraint learning framework}, which enables \methodName{} to learn to extract precise human motion in a self-supervised manner from any human sequences.

\begin{figure*}[t]
  \centering
  \includegraphics[width=\textwidth]{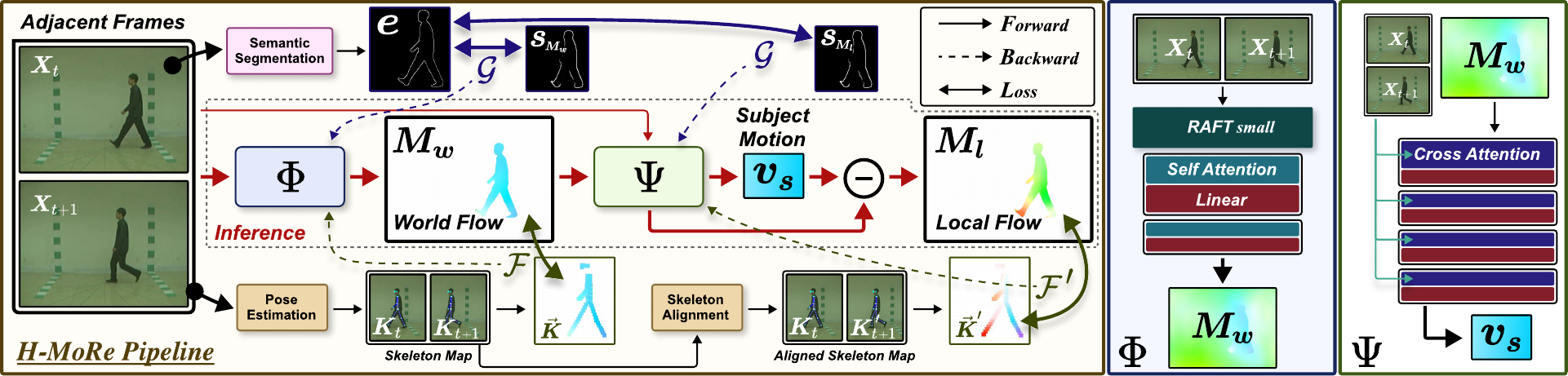}
  \caption{\mcaption{Whole Pipeline of \methodName{}.} \msubcaption{From left to right: (a) The training and inference pipeline for \emph{world-local} flows (outlined by a gray dashed line) and the use of the \emph{joint constraints learning framework} for self-supervised learning from real-world scenarios. Blue symbols and lines denote the boundary constraint \( \mathcal{G} \) , while green symbols and lines indicate the skeleton constraint \( \mathcal{F} \). Backpropagation gradients are shown as dashed lines in corresponding colors; (b) the internal implementation of \( \Phi \); and (c) the internal implementation of \( \Psi \).}}
  \label{fig:whole_pipeline}
  \mvspace
\end{figure*}

\subsection{Human-centric Motion Representation}
Researchers have explored various approaches to describing human motion for improving the performance of downstream tasks such as action recognition.
As an explicit motion representation, optical flow is widely used as an additional input channel to capture motion information~\cite{chen2023atm, carreira2017quo}.
However, as action analysis tasks have advanced, the complexity of human motion makes it increasingly difficult to extract useful information from optical flow.
Since optical flow is a non-human-oriented representation, it struggles to meet the specific demands of human motion analysis.
Consequently, many analysis systems turn to extract implicit human motion from pose information.
They employ GNNs to analyze the relationship between skeletal keypoints in each frame~\cite{yan2018spatial, teepe2021gaitgraph}. 
Early on, these skeletal keypoints are commonly detected by CNNs~\cite{he2017mask, bertasius2019learning, xu2022vitpose}. 
They represent the earliest human-centric motion representations.
With advancements in 3D computer vision, models like VideoPose3D~\cite{pavllo20193d} and MotionBERT~\cite{zhu2023motionbert} leverage temporal cues to improve the accuracy of pose estimation at each moment.
However, although pose information is sufficiently accurate, it overlooks the role of shape information in analysis tasks.
For example, gait recognition relies on the motion of specific body parts to identify a person, and the absence of the human shape makes pattern matching challenging.
Therefore, it is important to represent complex human motion while preserving the integrity of the human shape.
Motivated by this challenge, we introduce \emph{world-local flows}, which extract motion information across two distinct reference systems, providing a clear delineation of human boundaries and thereby enhancing the performance of various downstream tasks.
\section{Methodology {\small \mfinish{Nov.10} \mstatus{Ready}}}

\subsection{Method Overview} \label{sec:problem_formulation}


Motion representation is commonly computed by estimating optical flow. However, the estimated flow map represents both background and foreground motion, with human motion often obscured in the noisy map and thus challenging to use for downstream human action analysis tasks. In this paper, we propose a human-centric motion representation and demonstrate its benefits for these tasks. A complete list of symbols is provided in the Appendix (\cref{tab:vars_matrix}).

As shown in \cref{fig:whole_pipeline}, our method learns human-centric motion representations from perspectives: human motion relative to the environment, termed world flow \( \boldsymbol{M_w} \), and motion relative to the human subject, termed local flow \( \boldsymbol{M_l} \). Together, these \emph{world-local} provide comprehensive motion information for downstream tasks.

To begin, given two consecutive frames \(\boldsymbol{X}_{t} \) and \(\boldsymbol{X}_{t+1}\), we utilize a network \( \Phi \) to calculate a flow map as \( \boldsymbol{M_w} = \Phi \left( \boldsymbol{X}_{t}, \boldsymbol{X}_{t+1}; \, \omega_{1} \right) \), where \( \omega_{1} \) denotes the network parameters.
To obtain the optimal world flow \( \boldsymbol{M_w} \) with precise motion and detailed shape information, \ie, to optimize \( \omega_{1} \), we propose a \emph{joint constraint learning framework} that learns from real-world scenarios in a self-supervised manner.
The proposed learning framework has two fidelity terms: the skeleton constraint \( \mathcal{F} \) and the boundary constraint \( \mathcal{G} \).
The skeleton constraint \( \mathcal{F} \) uses pose information to ensure that each body point's movement adheres to kinematic constraints.
Specifically, \( \mathcal{F} \) quantifies the discrepancy between the estimated flow map \( \boldsymbol{M} \) and the skeleton offset between the poses in frames \(\boldsymbol{X}_{t}\) and \(\boldsymbol{X}_{t+1}\). 
To further incorporate body shape and fine-grained motion, the boundary constraint \( \mathcal{G} \) aligns human shapes onto the flow map by measuring the Chamfer distance between human boundaries in estimated flow \( \boldsymbol{M} \) and frame \( \boldsymbol{X}_{t} \).
Our world flow \( \boldsymbol{M_w} \) is then defined as:
\begin{equation}
    \boldsymbol{M_w} =  \arg\min_{\boldsymbol{M}}\left [ \mathcal{F}\left ( \boldsymbol{M}, \boldsymbol{X}_{t}, \boldsymbol{X}_{t+1} \right ) + \alpha \cdot \mathcal{G}\left ( \boldsymbol{M}, \boldsymbol{X}_{t}\right ) \right ], \label{eq:joint_constraint}
\end{equation}
where \( \alpha \) balances the contributions of each term. Details of \( \mathcal{F} \) and \( \mathcal{G} \) are discussed in \cref{sec:joint_constraint}.

\begin{figure}[t]
  \centering
  \includegraphics[width=0.48\textwidth]{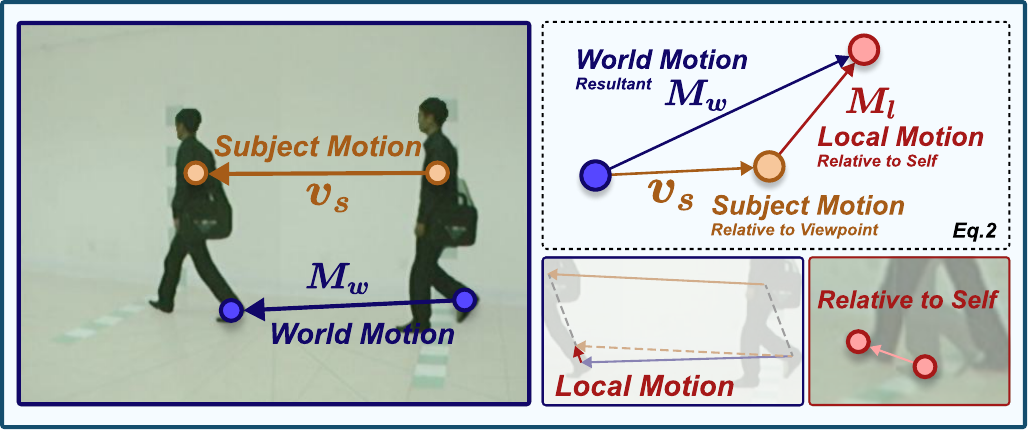}
  \caption{\mcaption{Definition of world-local flows.} \msubcaption{World motion \( \boldsymbol{M_w} \) is movement relative to the environment, while local motion \( \boldsymbol{M_l} \) is relative to the subject. Using the subject's overall motion \( \boldsymbol{v_s} \), these can be converted via vector composition and decomposition.}}
  \label{fig:wl_flows}
  \mvspace
\end{figure}

Although world flow \( \boldsymbol{M_w} \) is effective, certain tasks 
require a focus on motion relative to the subject.
In these scenarios, subject-relative motion (\eg, the arm's movement relative to the body) becomes more critical than motion relative to the environment.
We refer to this subject-relative motion as local flow \( \boldsymbol{M_l} \).

\begin{figure}[t]
  \centering
  \includegraphics[width=0.47\textwidth]{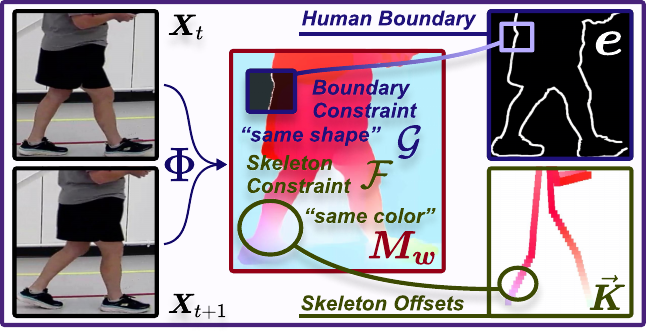}
  \caption{\mcaption{Overview of joint constraint learning framework.} \msubcaption{We introduce two constraints -- (1) the boundary constraint \( \mathcal{G} \) (blue), which aligns the human boundary with the flow edges to maintain consistent shapes; and (2) the skeleton constraint \( \mathcal{F} \), which uses skeleton offsets to regulate body point movements, ensuring consistent motion, as reflected by matching colors in the visualization.}} 
  \label{fig:joint_constraint}
  \mvspace
\end{figure} 

The estimation of local flow \( \boldsymbol{M_l} \) can theoretically be achieved using the same network architecture as that used for world flow \( \boldsymbol{M_w} \). However, this would double the inference time for \methodName{}, violating the requirement for real-time downstream tasks.
Inspired by Galilean transformations~\cite{galilei2002discourses} in kinematics, we identify a connection between world flow \( \boldsymbol{M_w} \) (blue vector) and local flow \( \boldsymbol{M_l} \) (red vector), as shown in \cref{fig:wl_flows}.
This allows us to avoid using a separate network \( \Phi' \) to estimate \( \boldsymbol{M_l} \). Instead, we employ a lightweight network \( \Psi \) to estimate the subject's overall motion trend \( \boldsymbol{v_s} \) (brown vector) based on \( \boldsymbol{M_w} \).
By applying vector decomposition, as shown in \cref{fig:whole_pipeline}, we can estimate local flow \( \boldsymbol{M_l} \) indirectly with lower computational cost:
\begin{equation}
    \begin{gathered}
        \boldsymbol{v_s} = \Psi \left(\boldsymbol{X}_{t}, \boldsymbol{X}_{t+1}, \boldsymbol{M_w}; \, \omega_{2} \right), \\
        \boldsymbol{M_l} = \boldsymbol{M_w} - \boldsymbol{v_s},
    \end{gathered}
    \label{eq:wl_flow}
\end{equation}
where \( \omega_{2} \) denotes the network parameters. Our joint constraint learning framework is similarly applied to \( \boldsymbol{M_l} \), as detailed in \cref{sec:wl_flow}.

\subsection{Joint Constraint Learning Framework} \label{sec:joint_constraint}

\begin{figure*}[t]
  \centering
  \includegraphics[width=\textwidth]{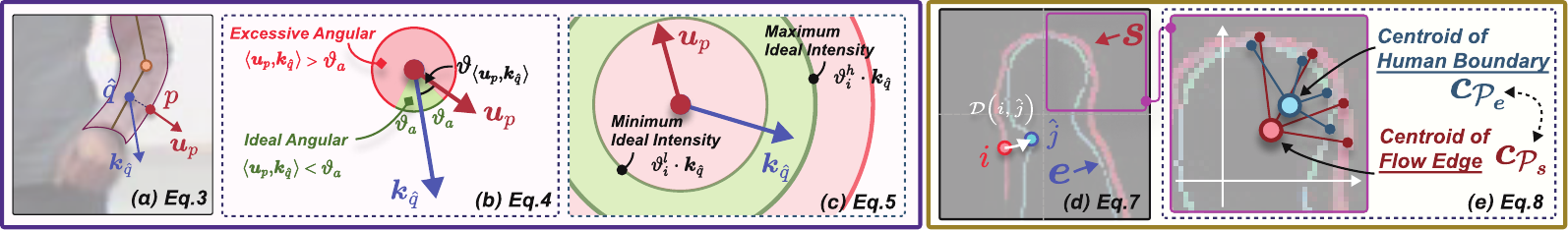}
  \caption{\mcaption{Details of our joint constraint.} \msubcaption{(a) matching any body point \(p\) to its corresponding skeleton point \(\hat{q}\); (b) an angular constraint to align estimated motion with skeleton's movement directions; (c) an intensity constraint to ensure consistent motion magnitude between estimated motion and skeleton offsets; (d) matching any point \(i\) on the flow edges \( \boldsymbol{s} \) to its corresponding \( \hat{j} \) on human boundaries \( \boldsymbol{e} \); and (e) calculation of our patch-centroid distance.}}
  \label{fig:skeletal_boundary}
  \mvspace
\end{figure*}


We discuss the skeleton constraint \( \mathcal{F} \) and the boundary constraint \( \mathcal{G} \) in \cref{eq:joint_constraint} for world flow \( \boldsymbol{M_w} \) in the following, and adapt them to local flow \( \boldsymbol{M_l} \) in \cref{sec:wl_flow}.

\subsubsection{Skeleton Constraint}
To define the overall range of motion for \methodName{}, we use pose information to establish a dynamic relationship between the skeleton and body.

Using 2D pose estimation (\eg, ED-Pose~\cite{yang2022explicit}), we extract \(17\) joint points from frames \( \boldsymbol{X}_t \) and \( \boldsymbol{X}_{t+1} \), each with corresponding coordinates and visibility \(c\).
By uniformly sampling joint-related \( n_{k} \) points along each bone segment, we interpolate these joint points to construct a skeleton map \( \boldsymbol{K}_{t} \) and subsequently calculate the skeleton offsets \( \vec{\boldsymbol{K}} = \boldsymbol{K}_{t+1} - \boldsymbol{K}_{t}\) (as shown in \cref{fig:whole_pipeline} and \cref{fig:joint_constraint}).

To constrain body motion using skeletal offsets, as shown in ~\cref{fig:skeletal_boundary}~(a), for any body point \( p \) (red point), we identify the closest skeletal keypoint \( \hat{q} \) (blue point) with the highest visibility.
This matching process is defined as:
\begin{equation}
    \hat{q} = \arg\min_{q \in \boldsymbol{K}_{t} \odot S} \left( - \left\| p - q \right\| \times \log{{c}_{q}} \right), ~\ \text{s.t.} \, p \in \boldsymbol{M} \odot S, 
    \label{eq:skeleton_match}
\end{equation}
where \( c_{q} \) is the visibility of \( q \). The \( \odot S \) operation restricts matching to the subject region \( S \), preventing incorrect voxel matches in flow \( \boldsymbol{M} \) when multiple subjects are present. 

Given the matched body-skeleton pair \( \langle p, \hat{q} \rangle \), we apply the skeletal offset at \( \hat{q} \) in \( \vec{\boldsymbol{K}} \), denoted as \( \boldsymbol{k}_{\hat{q}} \) (blue vector), to constrain the estimated flow \( \boldsymbol{M} \) at \( p \), denoted as \( \boldsymbol{u}_p \) (red vector).
Guided by the kinematic constraints of the human body, we assume that \( \boldsymbol{u}_p \) should be neither orthogonal to \( \boldsymbol{k}_{\hat{q}} \) nor significantly different in motion intensity. 
To enforce these properties, we define the skeleton constraint \( \mathcal{F} \) with two sub-constraints: an angular constraint \( \mathcal{F}_{A} \) and an intensity constraint \( \mathcal{F}_{I} \).

As shown in ~\cref{fig:skeletal_boundary}~(b), we define the angular constraint \( \mathcal{F}_{A} \) by calculating the angle between \( \boldsymbol{u}_p \) and \( \boldsymbol{k}_{\hat{q}} \).
We establish ideal (green sector) and excessive (red sector) motion ranges based on a empirically specified threshold \( \vartheta_{a} \), as:
\begin{equation}
    \mathcal{F}_{A}\left ( \boldsymbol{u}_p, \boldsymbol{k}_{\hat{q}} \right ) = \mathbb{I}\left ( \frac{\boldsymbol{u}_p\cdot \boldsymbol{k}_{\hat{q}}}{\left \| \boldsymbol{u}_p \right \| \cdot \left \| \boldsymbol{k}_{\hat{q}} \right \| } \geq \vartheta_{a} \right ),
    \label{eq:angular_constraint}
\end{equation}
where \( \mathbb{I} \left( \cdot \right) \) is an indicator function that identifies body points with abnormal motion.

Similarly, we expect the motion intensity of a body point to be consistent with that of its corresponding skeletal point.
As shown in ~\cref{fig:skeletal_boundary}~(c), we define the intensity constraint \( \mathcal{F}_{I} \) by establishing an acceptable intensity range (green ring) and error ranges for low or high intensities (red circle and ring). When \( \boldsymbol{u}_p \) falls outside the green ring, a penalty is applied:
\begin{equation}
    \mathcal{F}_{I}\left( \boldsymbol{u}_p, \boldsymbol{k}_{\hat{q}} \right)
    = \operatorname{ReLU} \left[
    \scalebox{0.85}{$
    \left( \left\| \boldsymbol{u}_p \right\| - \vartheta_i^l \cdot \left\| \boldsymbol{k}_{\hat{q}} \right\| \right)
    \cdot
    \left( \left\| \boldsymbol{u}_p \right\| - \vartheta_i^h \cdot \left\| \boldsymbol{k}_{\hat{q}} \right\| \right)
    $}
    \right]
    \label{eq:intensity_constraint}
\end{equation}

By applying both angular and intensity sub-constraints, we bind each body point's motion to skeletal motion, precisely defining the feasible movement range for each point.
The complete skeleton constraint \( \mathcal{F} \) in \cref{eq:joint_constraint} is thus:
\begin{equation}
    \mathcal{F} = \frac{1}{h \cdot w} \sum_{p \in \boldsymbol{M}} \left[ \mathcal{F}_{A}\left ( \boldsymbol{u}_p, \boldsymbol{k}_{\hat{q}} \right ) + \beta \cdot \mathcal{F}_{I}\left ( \boldsymbol{u}_p, \boldsymbol{k}_{\hat{q}} \right ) \right ],
    \label{eq:skeletal_constraint}
\end{equation}
where \( \beta \) balances the angular and intensity sub-constraints.

\subsubsection{Boundary Constraint} \label{sec:boundary_constraint}
To refine motion details and incorporate shape information, we use human boundaries \( \boldsymbol{e} \) as a prior to ensure that the learned flow \( \boldsymbol{M} \) closely adheres to the human body.

As illustrated in ~\cref{fig:skeletal_boundary}~(d), we minimize the Chamfer distance~\cite{aoki2019pointnetlk, yew2020rpm} between the human boundaries \( \boldsymbol{e} \) and flow edges \( \boldsymbol{s} \).
Human boundaries are detected using a Canny operator on the semantic segmentation results (\eg, from U\(^2\)Net~\cite{qin2020u2}), while flow edges are obtained from our custom algorithms detailed in the appendix.
This distance is defined as the sum of the nearest distances from each point \( i \) (red point) on the flow edges \( \boldsymbol{s} \) (red curve) to \( \boldsymbol{e} \) (blue curve):
\begin{equation}
    \begin{gathered}
        \mathcal{C}\left( \boldsymbol{s}, \boldsymbol{e} \right) = \sum_{i \in \boldsymbol{s}} \mathcal{D}\left( i, \hat{j} \right), \,
        \text{s.t.} \, \hat{j} = \arg\min_{j \in \boldsymbol{e}} \mathcal{D}\left( i, j \right),
        \label{eq:chamfer_distance}
    \end{gathered}
\end{equation}
where \( \mathcal{D}\left( \cdot \right) \) is the Euclidean distance between two points.

Directly matching points between the two dense curves \( \boldsymbol{e} \) and \( \boldsymbol{s} \) is computationally expensive.
For efficiency, we propose \emph{patch-centroid distance}, which approximates Chamfer distance via matrix operations.
Specifically, we divide both \( \boldsymbol{e} \) and \( \boldsymbol{s} \) into patches at multiple scales by applying kernels of varying receptive fields as shown in \cref{fig:skeletal_boundary}~(e).
For each patch \( \mathcal{P} \), the Chamfer distance between curves within that patch (\(\mathcal{P}_{\boldsymbol{s}}\) and \(\mathcal{P}_{\boldsymbol{e}}\)) is approximated by calculating the distance between their centroids \( \boldsymbol{c} \) (large red and blue points), assuming a smooth distribution of points:
\begin{equation}
        \mathcal{C}\left( \mathcal{P}_{\boldsymbol{s}}, \mathcal{P}_{\boldsymbol{e}} \right) = \frac{1}{n_p}\sum_{ { \left \{ \langle i, \hat{j} \rangle  \right \}  }  }\mathcal{D}\left ( i, \hat{j} \right ) 
        \approx \mathcal{D}\left ( \boldsymbol{c}_{\mathcal{P}_{\boldsymbol{s}}}, \boldsymbol{c}_{\mathcal{P}_{\boldsymbol{e}}} \right ),
        \label{eq:patch_centroid_distance}
\end{equation}
where \(n_p\) is the total number of point pairs in each patch, while the \( { \left \{ \langle i, \hat{j} \rangle  \right \}  } \) represents all matching pairs of \( \langle i, \hat{j} \rangle \).
The derivation of this approximation and its experimental validation are provided in the appendix.

To improve the accuracy of the approximation, we further divide patches and calculate patch-centroid distances at multiple scales. Thus, the boundary constraint \( \mathcal{G} \) is:
\begin{equation}
    \mathcal{G} = \frac{1}{n_{ms}}\sum_{ms}\left [ \frac{1}{n_{\mathcal{P}}} \sum_{\mathcal{P}\propto ms} \mathcal{C}\left( \mathcal{P}_{\boldsymbol{s}}, \mathcal{P}_{\boldsymbol{e}} \right)  \right ],
    \label{eq:edge_chamfer_constraint}
\end{equation}
where \(n_{ms}\) is the number of scales and \(n_{\mathcal{P}}\) is the number of patches at scale \(ms\).

\subsection{World-Local Flow Estimation} \label{sec:wl_flow}
Given the definitions of world-local flows in \cref{sec:problem_formulation}, \methodName{} operates in two stages.
First, we directly estimate world flow \( \boldsymbol{M_w} \). Then, we indirectly compute local flow \( \boldsymbol{M_l} \) by estimating subject motion \( \boldsymbol{v_s} \): $\boldsymbol{M_l} = \boldsymbol{M_w} - \boldsymbol{v_s}$ from Eq.~(\ref{eq:wl_flow}).

We estimate world flow \( \boldsymbol{M_w} \) between frames using flow estimation network \( \Phi \), whose backbone could be any optical flow estimation networks, \eg, RAFT~\cite{sui2022craft} or FlowFormer~\cite{huang2022flowformer}.
To reduce carbon emissions during training \methodName{} from scratch, we use RAFT-small as the backbone in \( \Phi \) in our experiments.
Additionally, as shown in \cref{fig:whole_pipeline}, we add two self-attention blocks to emphasize human-centric motion and filter out irrelevant information to motion analysis.
In training, the joint constraints in \cref{sec:joint_constraint} guide the optimization of \(\omega_{1}\) in \( \Phi \).

As described in \cref{eq:wl_flow}, we use a lightweight network \( \Psi \) to estimate subject motion \( \boldsymbol{v_s} \) based on world flow \( \boldsymbol{M_w} \) and input frames.
Our implementation of \( \Psi \) includes cross-attention blocks with a depth of \(4\).

During training, we use \cref{eq:joint_constraint} to optimize the parameters \( \omega_{1} \) in \( \Phi \) to obtain the world flow \( \boldsymbol{M}_w \).
However, since the skeleton offset at point \( \hat{q} \) (\( \boldsymbol{k}_{\hat{q}} \)) in \( \mathcal{F} \) (\cref{eq:skeletal_constraint}) represents movement relative to the environment, this constraint cannot be directly applied to the learning process of the local flow \( \boldsymbol{M}_l \), \ie, it cannot be used to optimize \( \omega_{2} \) in \( \Psi \).

To address this issue, we adjust the skeleton constraint (\( \mathcal{F} \)). Specifically, before computing the difference between the two skeleton maps \( \boldsymbol{K}_{t} \) and \( \boldsymbol{K}_{t+1} \), we first align \( \boldsymbol{K}_{t+1} \) to \( \boldsymbol{K}_{t} \), termed \( \boldsymbol{K}_{t+1}' \). This alignment can be achieved in various ways, \eg, solving a homography matrix by analytical solution or using head-based alignment. Consequently, we obtain \( \vec{\boldsymbol{K}}' \), and then replacing \( \boldsymbol{k}_{\hat{q}} \) with \( \boldsymbol{k}_{\hat{q}}' \) in \cref{eq:angular_constraint} and \cref{eq:intensity_constraint}. 
This results in a subject-relative constraint \( \mathcal{F}' \). Through this, we can determine the optimal \( \omega_{2} \) in \cref{eq:wl_flow}.

\section{Experiments {\small \mfinish{Nov.11} \mstatus{Updating}}} \label{sec:experiments}

\begin{table*}[t]
  \centering
  \centering
\resizebox{\textwidth}{!}{
    \small
    \begin{tabular}{lcccccccc} 
    \toprule
    \multirow{2}{*}{\textbf{Methods}}
    & \multicolumn{2}{c}{\textbf{Efficiency Analysis}}
    & \multicolumn{3}{c}{\textbf{GaitBase {\footnotesize [Params: 34.2M]}}}         
    & \multicolumn{3}{c}{\textbf{GaitSet {\footnotesize [Params: 6.3M]}}}
    \\
    \cmidrule(lr){2-3} \cmidrule(lr){4-6} \cmidrule(lr){7-9}
    & {\small\textbf{Params (M)}} & {\small\textbf{FLOPs (G)}}
    & {\small\textbf{NM@R1}\(\uparrow\)}    & {\small\textbf{BG@R1}\(\uparrow\)}    & {\small\textbf{CL@R1}\(\uparrow\)}
    & {\small\textbf{NM@R1}\(\uparrow\)}    & {\small\textbf{BG@R1}\(\uparrow\)}    & {\small\textbf{CL@R1}\(\uparrow\)}
    \\
    \midrule
    \text{w/o Flow} & - & -  & 96.51     & 91.50     & 78.02     & 92.82     & 82.25     & 69.24 \\
    \midrule
    RAFT \scriptsize{~\cite{teed2020raft}}                                 & 5.25     & 1780.4   & 96.91              & 93.12             & 80.52             & 93.28             & 82.21              & 70.55 \\
    GMA \scriptsize{~\cite{jiang2021learning}}                             & 5.88     & 2450.3   & 96.61              & 92.83             & 83.63             & \uline{93.72}     & 83.33              & 71.64 \\
    GMFlow \scriptsize{~\cite{xu2022gmflow}}                               & 4.68     & 428.6    & 97.72              & 93.59             & 85.22             & 93.62             & \uline{83.82}      & \uline{72.75} \\
    CRAFT \scriptsize{~\cite{sui2022craft}}                                & 6.30     & 4204.7   & 96.27              & 92.60             & 79.61             & 93.71             & 82.89              & 71.12 \\
    SKFlow \scriptsize{~\cite{sun2022skflow}}                              & 6.27     & 2731.5   & 96.59              & 92.42             & 82.13             & 93.52             & 82.38              & 70.53 \\
    VideoFlow \scriptsize{~\cite{shi2023videoflow}}                        & 12.65    & 3159.3   & \uline{97.88}      & 94.01             & \uline{86.12}     & 93.55             & 82.59              & 70.73 \\ 
    FlowFormer\scriptsize{++} \scriptsize{~\cite{shi2023flowformer++}}     & 16.15    & 3048.1   & 96.66              & \uline{94.31}     & 85.70             & 93.65             & 83.35              & 71.75 \\ 
    \midrule
    \textbf{\methodName{} \scriptsize{Ours}}                                        & 5.57    &   861.5   & \textbf{98.26}    & \textbf{95.62}    & \textbf{87.66}    & \textbf{98.27} & \textbf{94.46} & \textbf{85.25} \\
    \bottomrule
    \end{tabular}
}
  \caption{\mcaption{Quantitative comparison for gait recognition.} \msubcaption{Using GaitBase~\cite{fan2023opengait} and GaitSet~\cite{chao2021gaitset} (for real-time scenarios) as gait recognition models, we compare \methodName{} against seven optical flow estimation methods on the CASIA-B dataset. Higher values indicate better performance across all metrics. The best result is highlighted in \textbf{bold}, and the second-best is \uline{underlined}. The left two columns present an efficiency comparison of these flow estimation algorithms using the total parameters and FLOPs.}}
  \label{tab:gait_quantitative}
  \mvspace
\end{table*}

To evaluate the benefits of using \methodName{} as a motion representation for downstream tasks, we select three representative human-centric motion related tasks: gait recognition, action recognition, and video generation.
Our evaluation is conducted on three datasets: CASIA-B~\cite{zheng2011robust}, Diving48~\cite{li2018resound}, and UTD-MHAD~\cite{chen2015utd}. 
From each dataset’s training set, we sample \(2,800\), \(14,000\), and \(600\) sequences, respectively, to train \methodName{}. 
For each sequence, we extract \(16\) frames (\(8\) frames for Diving48) and resize them to \( 512 \times 384 \).

We set the tuning parameters \( \alpha \) and \( \beta \) to \(0.1\) and \(0.01\), and define the thresholds \(\vartheta_{a}\), \(\vartheta_{i}^{l}\), and \(\vartheta_{i}^{h}\) as \(15^\circ\), \(0.8\), and \(1.2\), respectively.
All learnable parameters are optimized using AdamW optimizer with a learning rate of \(1.0 \times 10^{-4}\) and exponential decay.
Training is conducted over \(8\) epochs with a batch size of \(64\). Our approach is implemented in PyTorch and runs on \(16\) NVIDIA RTX 6000 Ada GPUs.


\subsection{Results on Gait Recognition} \label{sec:gait_recognition}
Gait recognition is the task of identifying individuals based on unique walking patterns, where human motion is an important cue.
The CASIA-B dataset includes videos of \(124\) individuals recorded from \(11\) viewpoints under three conditions: normal walking (NW), bag carrying (BG), and clothing change (CL). 
The primary challenge in this task is to accurately recognize individuals despite variations in appearance, making precise motion information essential.

Rank-1 (R1) accuracy measures the percentage of times the model's top prediction matches the correct label.
We use GaitBase~\cite{fan2023opengait} as the gait recognition model to evaluate \methodName{}'s ability to integrate with complex downstream networks.
The inputs to GaitBase are per-frame silhouettes and flow (optical flow or \methodName{}) between consecutive frames.

As shown in \cref{tab:gait_quantitative},
under both NM and BG conditions, the model using \methodName{} as the motion representation achieves higher R1 accuracy compared to models using optical flow.
The CL condition is the most challenging in gait recognition, as clothing changes make it difficult to rely on body or clothing shape for identification.
In this case, accurate motion information is essential. The GaitBase model, leveraging motion information from \methodName{}, achieves top accuracy, showing a \(9.64\)\% improvement over the models that do not incorporate flow $(87.66\ \text{vs}\ 78.02)$.

Furthermore, to evaluate \methodName{} in real-time applications, we use a lightweight model, GaitSet~\cite{chao2021gaitset}.
As the motion representation, \methodName{} provides a \(12.50\)\% gain (CL@R1) over GaitSet using optical flow, underscoring \methodName{}'s advantage in real-time gait recognition.

\begin{table}[t]
  \centering
  \centering
\resizebox{0.48\textwidth}{!}{
    \small
    \begin{tabular}{lcccc} 
    \toprule
    \multirow{2}{*}{\textbf{Methods}}
    & \multicolumn{2}{c}{\textbf{Action Rec. \scriptsize{[Diving48]}}}
    & \multicolumn{2}{c}{\textbf{Video Gen. \scriptsize{[MHAD]}}}
    \\
    \cmidrule(lr){2-3} \cmidrule(lr){4-5}
    & {\small\textbf{Acc@1}\(\uparrow\)}    & {\small\textbf{Acc@5}\(\uparrow\)}
    & {\small \textbf{SSIM \(\uparrow\)}}   & {\small\textbf{FVD \(\downarrow\)}}
    \\
    \midrule
    \text{w/o Flow}              & 64.07             & 95.08              & 0.9463               & 329.22 \\
    \midrule
    RAFT                         & 65.90             & 95.42              & 0.9557               & 118.13 \\
    GMA                          & 69.00             & 95.62              & \uline{0.9565}       & \uline{114.98} \\
    GMFlow                       & \uline{71.74}     & 95.51              & 0.9529               & 429.82 \\
    CRAFT                        & 71.24             & 95.85              & 0.9532               & 424.72 \\
    SKFlow                       & 71.39             & 96.64              & 0.9541               & 325.08 \\
    VideoFlow                    & 71.45             & \uline{96.72}      & 0.9564               & 165.63 \\ 
    FlowFormer\scriptsize{++}    & 64.28             & 95.88              & 0.9553               & 284.95 \\ 
    \midrule
    \textbf{\methodName{} \scriptsize{Ours}}   & \textbf{72.99}    & \textbf{97.62} & \textbf{0.9574}    & \textbf{108.38} \\
    \bottomrule
    \end{tabular}
}
  \caption{\mcaption{Quantitative comparison for action recognition and video generation.} For action recognition, higher metric values indicate more accurate classification. In video generation, higher SSIM~\cite{wang2004image} scores reflect better frame quality, while lower FVD~\cite{unterthiner2018towards} scores indicate that the generated video more closely resembles real footage.}
  \label{tab:action_quantitative}
  \mvspace
\end{table}

\subsection{Results on Action Recognition} \label{sec:action_recognition}
Action recognition is widely used to classify human motion into categories, such as running and walking.
To evaluate \methodName{}'s capability in multi-person and high-speed scenarios, we test it on the Diving48 dataset, which contains approximately \(18,000\) sequences covering \(48\) types of diving actions. 
A primary challenge in Diving48 is that most dives occur within \(2\) seconds, often resulting in significant motion blur across frames.

We use Acc@1 and Acc@5 as metrics, which measure the percentage of times the model’s top-1 and top-5 predictions match the correct label, respectively.
Video-FocalNets~\cite{wasim2023video} is used as the action classifier, with frames and different flows (optical flow or \methodName{}) as inputs. These models are trained from scratch on the same dataset to ensure experimental fairness.

As shown in ~\cref{tab:action_quantitative}, the model using \methodName{} as the motion representation achieves a notable improvement in Acc@5 (\(0.9\)\% \(\uparrow\)), indicating enhanced robustness.
This improvement is attributed to our \emph{joint constraint learning framework}, which is specifically designed for human-centric scenarios, making \methodName{} less susceptible to challenges such as subject overlap, motion blur, and noisy backgrounds.
Additionally, we observe a substantial advantage in Acc@1, with an \(8.92\)\% improvement over the baseline model without any flow, due to \methodName{}'s precise motion information.

Given the real-time requirements of practical scenarios, such as sports events, we conduct experiments using TSN~\cite{wang2016temporal} as a more lightweight action classifier in place of Video-FocalNets.
In this setup, using our method as the motion representation yields a more pronounced improvement Acc@1 (\(1.67\)\%\(\uparrow\)).
Detailed results are in the appendix.


\begin{figure}[t]
  \centering
  \includegraphics[width=0.48\textwidth]{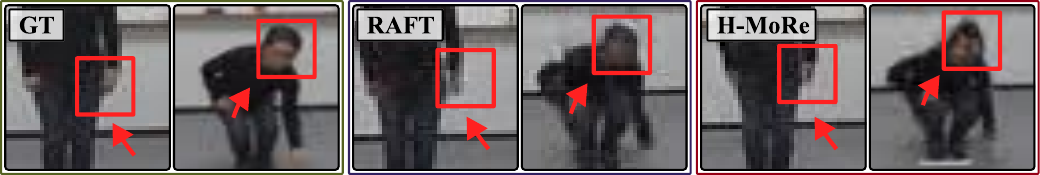}
  \caption{\mcaption{Qualitative comparison of video generation.} \msubcaption{Results of LGC-VD using RAFT and our \methodName{} as motion conditions. We show the generated first frame (\(1^{st}\)) and last frame (\(15^{th}\)) and corresponding ground truth (GT).}}
  \label{fig:video_generation}
  \mvspace
\end{figure}

\begin{figure*}[t]
  \centering
  \includegraphics[width=\textwidth]{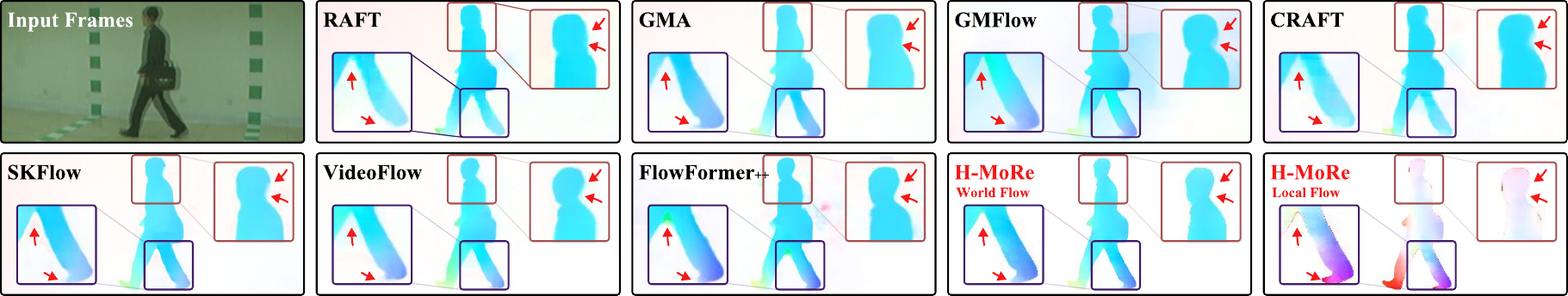}
  \caption{\mcaption{Qualitative comparison of flow visualizations.} \msubcaption{Flow visualizations generated by our \methodName{} and seven SoTA optical flow estimation algorithms on the same frame pair. Primary differences are marked and zoomed with red boxes and arrows.}}
  \label{fig:qualitative}
\end{figure*}

\subsection{Results on Video Generation}
Understanding motion is essential for effective video generation.
To access whether \methodName{} can provide accurate motion information for video generation models, we conduct experiments on the UTD-MHAD dataset. 
This dataset includes \(861\) action sequences spanning \(27\) action types performed by \(8\) subjects, with each action repeated \(4\) times.

We use SSIM~\cite{wang2004image} and FVD~\cite{unterthiner2018towards} as metrics. 
SSIM measures the spatial consistency between generated and real videos at the frame level, reflecting the quality of generated frames. 
FVD evaluates the consistency between generated and real videos in high-dimensional feature space, capturing the dynamic coherence and realism of the video.

We use LGC-VD~\cite{yang2023video} as the generation model, taking the first two frames as input and using flows from various methods as conditioning information. 
As shown in~\cref{tab:action_quantitative}, compared to generation models using optical flow, the model with \methodName{} achieves a higher SSIM (\(0.25\)\%\(\uparrow\)) and lower FVD (\(5.74\)\%\(\downarrow\)).
As illustrated in \cref{fig:video_generation}, in the first and last frames generated by models conditioned on RAFT, noticeable distortions or missing appear in hands and head (marked with red boxes).
This degradation occurs because optical flow inaccurately estimates human and environmental motion, resulting in unclear body shapes and causing generation models to struggle with retaining texture details.
In contrast, our \methodName{} is tailored for human-centric scenarios,  and ensures high generation quality.

\begin{figure}[t]
  \centering
  \includegraphics[width=0.48\textwidth]{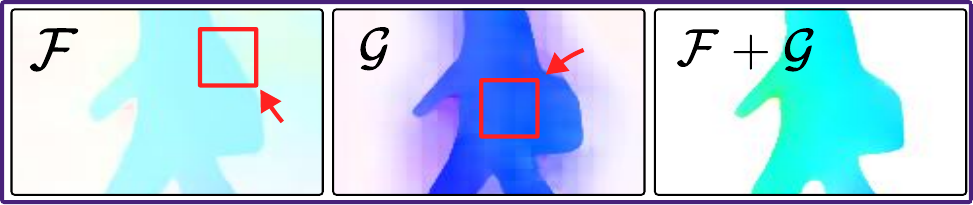}
  \caption{\mcaption{Ablation study on joint constraint.} \msubcaption{Visual comparison of motion estimation results using individual constraints versus joint constraints.}}
  \label{fig:constraint_ablation}
\end{figure}

\subsection{Qualitative Comparison}

In ~\cref{fig:qualitative}, we present a qualitative comparison of \methodName{}'s \emph{world-local flows} on the CASIA-B dataset, highlighting the distinctions between our method and optical flow estimation methods.
The zoom-in regions show \methodName{}'s advantages: thanks to the skeleton constraint, our method provides more accurate overall motion.
Additionally, the boundary constraints allow for precise human edge details in areas such as the head and feet.
Most importantly, the unique representation of local flow further underscores the underlying motion information, setting it apart from other methods.

\subsection{Ablation Studies}

\paragraph{Study on Joint Constraint} 
We ablate the effects of the two constraints in  \methodName{}. \cref{tab:constraint_ablation} and \cref{fig:constraint_ablation} show the results when each constraint is applied individually.
Using only the skeleton constraint \( \mathcal{F} \)  aligns motion direction and magnitude but introduces unintended motion information, as shown by the red box in \cref{fig:constraint_ablation}. 
In contrast, applying only the boundary constraint \( \mathcal{G} \) yields precise human edges but leads to substantial errors in motion information.
Combining two constraints results in a representation that accurately captures both precise motion and shape details, optimizing \methodName{} for human-centric motion representations.

\begin{table}[t]
  \centering
  \centering
\resizebox{0.48\textwidth}{!}{
    \small
    \begin{tabular}{>{\centering\arraybackslash}p{18pt} >{\centering\arraybackslash}p{18pt} cccccc}
    \toprule
    \multicolumn{2}{c}{\textbf{Constraint}}        
    & \multicolumn{3}{c}{\textbf{Gait Rec.}}
    & \multicolumn{2}{c}{\textbf{Action Rec.}}
    \\
    \cmidrule(lr){1-2} \cmidrule(lr){3-5} \cmidrule(lr){6-7}
    \textbf{\(\mathcal{F}\)} & \textbf{\(\mathcal{G}\)}
    & {\small\textbf{NM}{\tiny@R1}\(\uparrow\)} & {\small\textbf{BG}{\tiny@R1}\(\uparrow\)} & {\small\textbf{CL}{\tiny@R1}\(\uparrow\)}
    & {\small\textbf{Acc{\tiny@1}}\(\uparrow\)}    & {\small\textbf{Acc{\tiny@5}}\(\uparrow\)}
    \\
    \midrule
    \checkmark &                & \uline{97.15}     & 91.47             & 83.01             & \uline{72.13}     & \uline{96.14} \\
    & \checkmark                & 96.99             & \uline{93.09}     & \uline{84.93}     & 68.17     & 95.23 \\
    \midrule
    \checkmark & \checkmark     & \textbf{98.26}    & \textbf{95.62}    & \textbf{85.25}    & \textbf{72.99}     & \textbf{97.62} \\
    \bottomrule
    \end{tabular}
}
  \caption{\mcaption{Ablation study on joint constraint.} \msubcaption{Quantitative comparison results on gait recognition and action recognition.}}
  \label{tab:constraint_ablation}
\end{table}

\begin{table}[t]
  \centering
  \centering
\resizebox{0.48\textwidth}{!}{
    \small
    \begin{tabular}{>{\centering\arraybackslash}p{18pt} >{\centering\arraybackslash}p{18pt} cccccc}
    \toprule
    \multicolumn{2}{c}{\textbf{W-L Flows}}        
    & \multicolumn{3}{c}{\textbf{Gait Rec.}}
    & \multicolumn{2}{c}{\textbf{Action Rec.}}
    \\
    \cmidrule(lr){1-2} \cmidrule(lr){3-5} \cmidrule(lr){6-7}
    \textbf{\(\boldsymbol{M_w}\)} & \textbf{\(\boldsymbol{M_l}\)}
    & {\small\textbf{NM}{\tiny@R1}\(\uparrow\)} & {\small\textbf{BG}{\tiny@R1}\(\uparrow\)} & {\small\textbf{CL}{\tiny@R1}\(\uparrow\)}
    & {\small\textbf{Acc{\tiny@1}}\(\uparrow\)}    & {\small\textbf{Acc{\tiny@5}}\(\uparrow\)}
    \\
    \midrule
    \checkmark &                & 97.13             & 92.31             & 80.78             & 70.91              & \uline{95.79}    \\
    & \checkmark                & \uline{98.13}     & \uline{94.41}     & \uline{82.82}     & \uline{72.64}      & 95.38            \\
    \midrule
    \checkmark & \checkmark     & \textbf{98.26}    & \textbf{95.62}    & \textbf{85.25}    & \textbf{72.99}     & \textbf{97.62}   \\
    \bottomrule
    \end{tabular}
}
  \caption{\mcaption{Ablation study on world-local flows.} \msubcaption{Quantitative comparison results on gait recognition and action recognition.}}
  \label{tab:wlflow_ablation}
  \mvspace
\end{table}

\paragraph{Study on World-Local Flows}
To analyze the unique characteristics of our \emph{world-local flows}, we conduct experiments using only world flow \( \boldsymbol{M_w} \) or local flow \( \boldsymbol{M_l} \) on gait recognition and action recognition tasks.
As shown in \cref{tab:wlflow_ablation}, results indicate that using only local flow \( \boldsymbol{M_l} \) generally achieves higher accuracy than using world flow \( \boldsymbol{M_w} \) alone, supporting our assertion in \cref{sec:problem_formulation} that certain tasks rely more on motion relative to the subject than the environment.
More importantly, combining both flows, \ie, \emph{world-local flows}, yields higher accuracy than either world or local alone, highlighting their complementary nature.
This combination provides more insightful motion representations for human-centric action analysis, as demonstrated in \cref{tab:gait_quantitative} and \cref{tab:action_quantitative}.

\subsection{Comparison with Pose Estimation Methods}
As mentioned in \cref{sec:introduction}, optical flow and human pose are two common motion representations.
Since all baselines in \cref{tab:gait_quantitative} and \cref{tab:action_quantitative} are optical flow estimation methods, we now compare \methodName{} with models that use pose as the motion representation, including 2D pose from AlphaPose~\cite{fang2022alphapose} and 3D pose from MotionBERT~\cite{zhu2023motionbert}.
In the comparisons in previous subsections, all flows are used as additional input channels, concatenated with silhouettes (for gait recognition) or frames (for action recognition).
Because CNNs and ViTs cannot directly process pose information, in this subsection we use ST-GCN~\cite{yan2018spatial} to extract motion features from the pose data.
These features are then fused with visual features extracted by CNNs or ViTs through cross-attention layers and used as input to the classification head.

As shown in \cref{tab:pose_quantitative}, \methodName{} achieves significantly higher accuracy across all three settings in gait recognition, due to the additional shape information we provide.

\begin{table}[t]
  \centering
  \centering
\resizebox{0.48\textwidth}{!}{
    \small
    \begin{tabular}{lccccc}
    \toprule
    \multirow{2}{*}{\textbf{Representation}}
    & \multicolumn{3}{c}{\textbf{Gait Rec.}}
    & \multicolumn{2}{c}{\textbf{Action Rec.}}
    \\
    \cmidrule(lr){2-4} \cmidrule(lr){5-6}
    & {\small\textbf{NM}{\tiny@R1}\(\uparrow\)} & {\small\textbf{BG}{\tiny@R1}\(\uparrow\)} & {\small\textbf{CL}{\tiny@R1}\(\uparrow\)}
    & {\small\textbf{Acc{\tiny@1}}\(\uparrow\)}    & {\small\textbf{Acc{\tiny@5}}\(\uparrow\)}
    \\
    \midrule
    2D Pose \scriptsize{[\(\sim\)25M]}                             & 96.82             & 93.44             & 82.17             & 70.66              & 96.11 \\
    3D Pose \scriptsize{[\(\sim\)41M]}                             & \uline{97.21}     & \uline{94.32}     & \uline{84.05}     & \uline{71.51}      & \uline{97.09} \\
    \midrule
    \textbf{\methodName{} \scriptsize{[\(\sim\)5.7M]}}                       & \textbf{98.26}    & \textbf{95.62}    & \textbf{87.66}    & \textbf{72.99}     & \textbf{97.62} \\
    \bottomrule
    \end{tabular}
}
  \caption{\mcaption{Quantitative comparison with pose-based models.} \msubcaption{Performance of gait and action recognition using pose information versus \methodName{} as the motion representation.}}
  \label{tab:pose_quantitative}
  \mvspace
\end{table}


\subsection{Robustness for Overlapping Subjects}

\begin{figure}[t]
  \centering
  \includegraphics[width=0.48\textwidth]{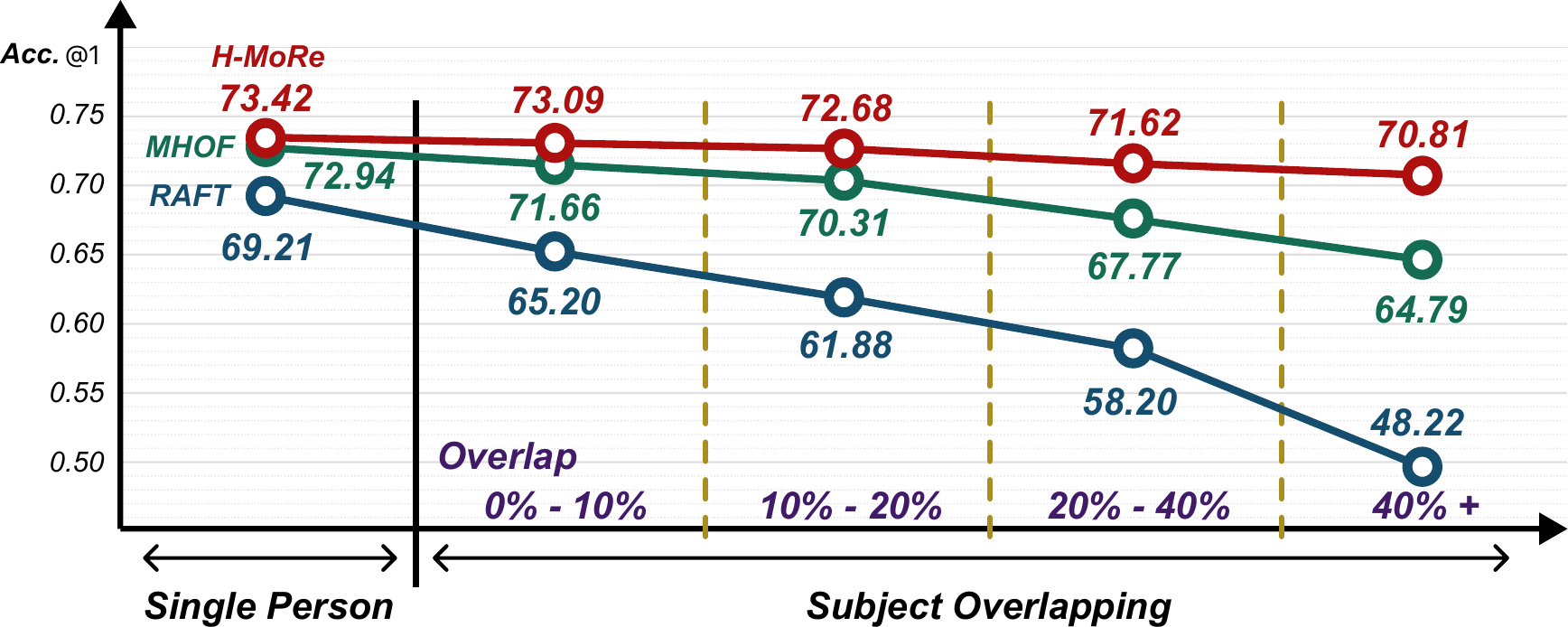}
  \caption{\mcaption{Robustness with increasing subject overlap.} \msubcaption{As subject overlap increases, action recognition model's accuracy using \methodName{} (red) decreases more slowly than others.}}
  \label{fig:overlap_robust}
  \mvspace
\end{figure}

We evaluate \methodName{}'s robustness in scenarios with overlapping subjects.
Based on the degree of bounding box overlap between subjects, we divide Diving48 into five bins, ranging from no overlap to \(40\)\% overlap, as shown in \cref{fig:overlap_robust}.
Since RAFT is not designed with a human-centric motivation, it struggles to distinguish pixels from multiple subjects. 
As overlap levels increase, the accuracy of classifiers using RAFT (blue line) as the motion representation declines rapidly.
We also compare against MHOF~\cite{ranjan2020learning}, which was trained under supervision on a synthetic multi-person dataset.
Due to its lack of exposure to real-world high-speed motion, classifiers using MHOF (green line) exhibit a noticeable performance drop once the subject overlap exceeds \(20\%\).
In contrast, \methodName{}, tailored for human-centric scenarios, mitigates the impact of increasing overlap, resulting in a significantly slower decline in accuracy.
\section{Conclusion {\small \mfinish{Nov.11} \mstatus{Ready}}}

In this paper, we introduce \methodName{}, a novel pipeline designed to enhance the precision of human-centric motion analysis.
Through a \emph{joint constraint learning framework} and motion representation across two distinct reference systems, \ie, \emph{world-local flows}, our method provides deeper insights into human motion. Experimental results show that our method substantially improves performance across diverse downstream tasks.
Due to computational constraints, our approach is primarily validated in 2D with a limited number of subjects. Future work should aim to extend this framework to 3D environments and more complex multi-subject scenarios, potentially increasing \methodName{}'s applicability and accuracy in real-world applications.

\section*{Acknowledgment}

This research is partially supported by the Army Research Office (ARO) grant W911NF-24-1-0385, the Office of Naval Research (ONR) grant N00014-23-1-2046, and the Office of the Director of National Intelligence (ODNI), Intelligence Advanced Research Projects Activity (IARPA), via 2022-21102100004. The views and conclusions contained in this document are those of the authors and should not be interpreted as necessarily representing the official policies, either expressed or implied, of ARO, ONR, ODNI, IARPA, or the U.S. Government. The U.S. Government is authorized to reproduce and distribute reprints for governmental purposes notwithstanding any copyright annotation therein.

{
    \small
    \bibliographystyle{ieeenat_fullname}
    \bibliography{refer/alias, refer/main}
}

\clearpage
\setcounter{page}{1}
\setcounter{section}{0}
\maketitlesupplementary

\renewcommand{\thesection}{\Alph{section}}

\section{Appendix}
\label{sec:appendix}

\subsection{Adaptive Edge Detection from Flow Maps}

In \cref{sec:boundary_constraint}, to refine motion details and incorporate shape information, we build boundary constraint based on human boundaries and edges of flow maps. The human boundaries are detected using a Canny operator.
However, the edges of the flow map cannot be extracted using simple operators like the Canny operator. Therefore, we designed an edge detection method with learnable thresholds specifically for flow maps.
We define the edges of a flow map \( \boldsymbol{M} \) as a series of discrete points, meaning that these points exhibit either intensity or angular discontinuities relative to their neighboring points. Intensity discontinuities indicate significant differences in offset magnitude between a point \(i\) and its neighbors \(j\), \eg, the boundary between a moving foreground and a static background. This can be mathematically expressed as:
\begin{equation}
    \boldsymbol{s}_{I} = \left \{ i \in \boldsymbol{M} \mid \left | \|\boldsymbol{M}_{i}\| - \|\boldsymbol{M}_{j}\| \right |  \ge \vartheta_{i} \right \},
\end{equation}
where \(\vartheta_{i}\) represents the learnable intensity threshold.
On the other hand, angular discontinuities refer to situations where the angle of the offset between \(i\) and \(j\) exhibits a significant difference, often occurring between different body patterns. This can be represented as:
\begin{equation}
    \boldsymbol{s}_{A} = \left \{ i \in \boldsymbol{M} \mid \frac{\boldsymbol{M}_{i}\cdot \boldsymbol{M}_{j}}{\| \boldsymbol{M}_{i} \| | \cdot \| \boldsymbol{M}_{j} \| } \geq \vartheta_{a} \right \},
\end{equation}
where \(\vartheta_{a}\) represents the learnable angular threshold.
Therefore, for any flow map \(\boldsymbol{M}\), its edge map \(\boldsymbol{s}\) can be formulated as the union of intensity and angular discontinuities:
\begin{equation}
    \boldsymbol{s} = \boldsymbol{s}_{I} \cup \boldsymbol{s}_{A}.
\end{equation}

\subsection{Patch-Centroid Distance Validation}

\begin{figure}[t]
  \centering
  \includegraphics[width=0.48\textwidth]{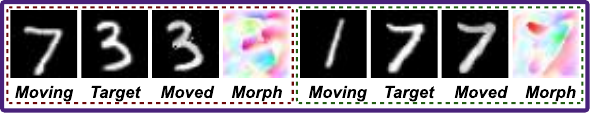}
  \caption{\mcaption{Validation for patch-centroid distance.} \msubcaption{By using the patch-centroid distance as the sole loss function, we can effectively train a network to deform one curve into another by applying an estimated morph.}}
  \label{fig:morph}
\end{figure}

In \cref{eq:patch_centroid_distance}, we propose a method to approximate the Chamfer distance using the patch-centroid distance. Here, we provide some validations for this approximation.
Based on the following formulation, the Chamfer distance can be approximately transformed, as shown below, under the condition of sufficient curve smoothness, leading to the theoretical conclusion presented in the main text:
\begin{equation}
    \begin{aligned}
        \mathcal{C}\left( \mathcal{P}_{\boldsymbol{s}}, \mathcal{P}_{\boldsymbol{e}} \right) &= \frac{1}{n_p}\sum_{ { \left \{ \langle i, \hat{j} \rangle  \right \}  }  }\mathcal{D}\left ( i, \hat{j} \right ) 
        \\ &\approx \mathcal{D}\left [ \frac{1}{n_p}\sum_{ { \left \{ \langle i, \hat{j} \rangle  \right \}  }  } i,  \frac{1}{n_p}\sum_{ { \left \{ \langle i, \hat{j} \rangle  \right \}  }  } \hat{j} \right ] 
        \\ &\approx \mathcal{D}\left ( \boldsymbol{c}_{\mathcal{P}_{\boldsymbol{s}}}, \boldsymbol{c}_{\mathcal{P}_{\boldsymbol{e}}} \right ).
    \end{aligned}
\end{equation}
Apart from the theoretical validation, we also conduct experiments on the MNIST dataset to verify whether the proposed patch-centroid distance can effectively measure the distance between two curves. As shown in \cref{fig:morph}, in the experiment, a network tries to apply a non-rigid transformation (Morph) to a moving image (Moving) to generate a moved image (Moved), aligning it with a specified target image (Target). By using the patch-centroid distance as the sole loss function to measure the curve distance between the moved and target images, the network successfully converges. This experimental result further demonstrates the patch-centroid distance as an effective approximation of the Chamfer distance.

\begin{figure}[t]
  \centering
  \includegraphics[width=0.48\textwidth]{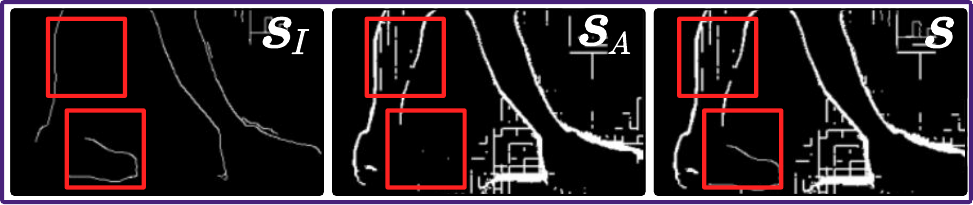}
  \caption{\mcaption{Edges of flow map.} \msubcaption{Through the complementarity of intensity and angular edges (highlighted in the red box), we can effectively detect the edges present in the flow map.}}
  \label{fig:flow_edges}
  \mvspace
\end{figure}

\subsection{Evaluating Action Recognition with TSN}

Similar to \cref{sec:gait_recognition}, besides the results shown in \cref{tab:action_quantitative}, we also validate the improvement in real-time motion analysis performance achieved by \methodName{} on the action recognition task.
Following the same experimental setup as described in \cref{sec:action_recognition}, we conducted a quantitative comparison on the Diving48 dataset using TSN instead of Video-FocalNets as the action recognition classifier.
To validate the improvement in real-time motion analysis performance achieved by \methodName{}, we conducted a quantitative comparison on the Diving48 dataset using TSN as the action recognition classifier. As shown in \cref{tab:quan_tsn}, compared to using optical flow as the motion representation input, using \methodName{} as the input significantly improved classification performance. This further demonstrates the effectiveness of \methodName{} in real-time scenarios. Besides, due to the use of additional output channels (optical flow:~2 more channels; \methodName{}:~4 more channels), the number of parameters in classifiers fluctuates slightly compared to the vanilla TSN. However, this does not affect performance. We have also indicated these fluctuations in the tables (Params).

\begin{table}[t]
  \centering
  \centering
\resizebox{0.48\textwidth}{!}{
    \small
    \begin{tabular}{lccc} 
    \toprule
    {\small\textbf{Methods}}
    & {\small\textbf{Acc@1}\(\uparrow\)}    & {\small\textbf{Acc@5}\(\uparrow\)} & {\small\textbf{Params (M)}}
    \\
    \midrule
    \text{w/o Flow}              & 65.58             & 95.18         & 4.5     \\
    \midrule
    RAFT                         & 66.09             & 93.45         & 4.6 + 5.25    \\
    GMA                          & 69.54             & 94.77         & 4.6 + 5.88    \\
    GMFlow                       & 70.91             & 95.89         & 4.6 + 4.68    \\
    CRAFT                        & 70.20             & 95.74         & 4.6 + 6.30    \\
    SKFlow                       & 67.26             & 94.57         & 4.6 + 6.27    \\
    VideoFlow                    & \uline{71.07}     & \uline{96.80} & 4.6 + 12.65   \\
    FlowFormer\scriptsize{++}    & 70.66             & 95.94         & 4.6 + 16.15   \\
    \midrule
    \textbf{\methodName{} \scriptsize{Ours}}   & \textbf{72.69}    & \textbf{97.60} & 4.7 + 5.57 \\
    \bottomrule
    \end{tabular}
}
  \caption{\mcaption{Quantitative comparison for action recognition in real-time scenarios.} \msubcaption{Alongside with numbers of learnable parameters in whole recognition pipeline containing motion estimation networks and classifiers.}}
  \label{tab:quan_tsn}
\end{table}

\subsection{Skeleton Map Alignment Methods}

In \cref{sec:wl_flow}, we mentioned that to impose a skeleton constraint on the local flow \(\boldsymbol{M_l}\), the original skeleton constraint \(\mathcal{F}\) needs to be transformed to \(\mathcal{F}'\).
This is because the \(\vec{\boldsymbol{K}}\) used in \(\mathcal{F}\) represents the skeleton's offset relative to the environment rather than relative to the subject itself. Therefore, we need to compute the skeleton offset relative to the subject itself by aligning the skeleton map \(\boldsymbol{K}_{t+1}\) from frame \(\boldsymbol{X}_{t+1}\) to \(\boldsymbol{K}_{t}\) from \(\boldsymbol{X}_{t}\). In practice, we applied two different alignment methods: (i) full-body and (ii) head-anchor.
Full-body alignment aligns \(K_{t+1}\) by solving the following equation using the least squares method:
\begin{equation}
    \begin{gathered}
            H' = \arg\min_{H}\left \| H \times \boldsymbol{K}_{t+1} - \boldsymbol{K}_{t} \right \|,
            \\ \boldsymbol{K}_{t+1}' = H' \times \boldsymbol{K}_{t+1}.
    \end{gathered}
\end{equation}
Here, \(H\) is a homography matrix, which enables skeleton map alignment through projection. This method is suitable for scenarios where the human body does not undergo rotation, such as in gait recognition or video generation.

However, when body rotation occurs, as in diving scenarios, full-body alignment based on all skeletal points may lead to errors in motion estimation. To address this, we use head-anchor alignment. This method employs an affine transform to rotate and scale \(\boldsymbol{K}_{t+1}\), ensuring the head regions in the skeleton maps at the two time points are closely matched. Based on this alignment, we obtain the transformed \(\boldsymbol{K}_{t+1}'\).


\subsection{Details of Experimental Settings}

In \cref{sec:experiments}, we compare the accuracy of \methodName{} in representing motion against optical flows across three tasks. In this subsection, we provide detailed explanations and additional information about our experimental workflow.

\subsubsection{Gait Recognition}

\begin{figure}[t]
  \centering
  \includegraphics[width=0.48\textwidth]{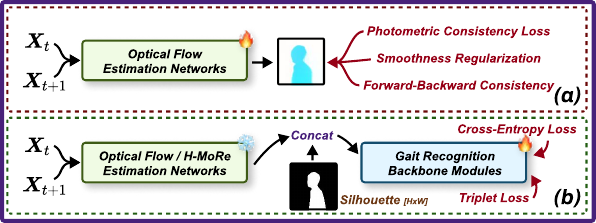}
  \caption{\mcaption{Pipeline for gait recognition experiments.} \msubcaption{(a) illustrates how we fine-tune the optical flow estimation networks. (b) demonstrates how motion information is integrated into the gait recognition pipeline.}}
  \label{fig:workflow_gait}
\end{figure}

For the gait recognition task, we first extract \(2,800\) sequences from the CASIA-B training set to fine-tune the optical flow estimation models (\cref{fig:workflow_gait}~(a)) and train our \methodName{} as denoted in \cref{sec:wl_flow}. After fine-tuning and training, the parameters of these motion estimation networks are frozen. We then use these models as inputs to train classifiers, specifically GaitBase or GaitSet networks with identical structures (\cref{fig:workflow_gait}~(b)).

Since silhouettes are the default input for gait recognition and require a single-channel input, we adjust the input layer of the classifier for optical flow (2 channels) to support three-channel input. For \methodName{}, which includes world flow (2 channels) and local flow (2 channels), we use a five-channel input. During this stage, only the classifier's parameters are trained. After training, all parameters are fixed, and the models are then tested on the testing set, producing the results shown in \cref{tab:gait_quantitative}.

\subsubsection{Action Recognition}

For the action recognition task, we similarly extract \(14,000\) sequences from the Diving48 training set to fine-tune the optical flow estimation models and train \methodName{} (\cref{fig:workflow_gait}~(a)). After freezing the parameters of these motion estimation networks, we train downstream classifiers: Video-FocalNets (\cref{tab:action_quantitative}) and TSN (\cref{tab:quan_tsn}).

For each moment, RGB images, which are the default input for action recognition, are combined with optical flow or \methodName{} and fed into the classifier. Specifically, these inputs are either (RGB 3 channels + Optical Flow 2 channels) or (RGB 3 channels + \methodName{} 4 channels).

\begin{figure}[t]
  \centering
  \includegraphics[width=0.48\textwidth]{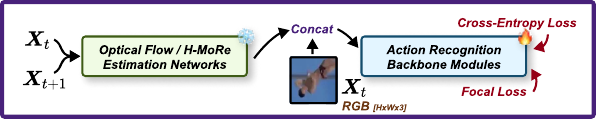}
  \caption{\mcaption{Pipeline for action recognition experiments.} \msubcaption{Similar to the gait recognition pipeline, the main difference lies in using RGB frames as additional input instead of silhouettes.}}
  \label{fig:workflow_action}
\end{figure}

\subsubsection{Video Generation}

Our video generation tasks in the main text can be regarded as motion-guided video reconstruction tasks. Unlike the first two tasks, this task is designed to directly evaluate the accuracy of motion provided by \methodName{}. Here, we use the motion at each time step as a condition for a diffusion model to reconstruct the original video. This implies that ignoring lighting variations, the SSIM and FVD scores in \cref{tab:action_quantitative} are highly correlated with the accuracy of the motion information provided by optical flow and \methodName{}.

\begin{figure}[t]
  \centering
  \includegraphics[width=0.48\textwidth]{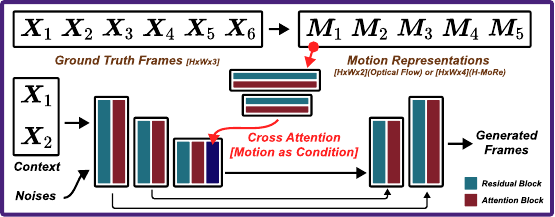}
  \caption{\mcaption{Pipeline for video generation experiments.} \msubcaption{This figure illustrates how we use motion representations as conditions for the video generation model to guide video reconstruction. For more details on the network structure, please refer to the LGC-VD.}}
  \label{fig:workflow_video}
\end{figure}

We fine-tune the optical flow estimation models and train \methodName{} using \(600\) sequences from the UTD-MHAD dataset, then freeze their parameters. 
Notably, unlike the pipelines in the previous tasks, this task does not concatenate motion directly with input (the first two frames of the sequence). Instead, the motion is used as a condition and integrated into the diffusion models via cross-attention (\cref{fig:workflow_video}).

\vspace{8pt}
\noindent These three tasks collectively demonstrate the effectiveness of \methodName{} in providing accurate motion information, either indirectly (gait recognition and action recognition) or directly (video generation). Additionally, \cref{fig:qualitative} in the main text and attached video provide a more intuitive visual comparison between our \methodName{} and optical flow, highlighting their respective differences and characteristics.

\section{Variables}

In the following table, we summarize the symbols used in the main text, along with their detailed definitions and representations.

\begin{table}[htbp]
    \centering
    \newcommand{\mtmain}[1]{\textbf{#1}}
\newcommand{\mtsub}[1]{{#1}} 
\newcommand{\mtcell}[1]{\makecell[lt]{#1}}

\resizebox{0.48\textwidth}{!}{
    \small
    \begin{tabular}{lp{5.5cm}l}
    \toprule
    \textbf{Variables} & \textbf{Description} & \textbf{Type {\footnotesize[shape]}} \\
    \midrule
    \(\boldsymbol{M_w}\) & \mtmain{\methodName{}'s world flow.} \mtsub{It represents the offset of each body point relative to the environment, \eg, the ground or the camera.} & \mtcell{Matrix \\ {\footnotesize\(\left[ H \times W \times 2 \right]\)}} \\ 
    \(\boldsymbol{M_l}\) & \mtmain{\methodName{}'s local flow.} \mtsub{It represents the offset of each body point relative to the subject themselves.} & \mtcell{Matrix \\ {\footnotesize\(\left[ H \times W \times 2 \right]\)}} \\ 
    \(\boldsymbol{X}_{t}\) \(\boldsymbol{X}_{t+1}\) & \mtmain{Two consecutive frames.} \mtsub{They are usually two frames from a video, spaced more than 0.1 seconds apart.} & \mtcell{Matrix \\ {\footnotesize\(\left[ H \times W \times 3 \right]\)}} \\ 
    \(\boldsymbol{K}_{t}\) \(\boldsymbol{K}_{t+1}\) & \mtmain{Skeleton maps.} \mtsub{They represent a series of (210) skeletal points of a person in frames \(\boldsymbol{X}_{t}\) and \(\boldsymbol{X}_{t+1}\), with each point containing coordinates and visibility information (\(c\)).} & \mtcell{Matrix \\ {\footnotesize\(\left[ 210 \times 3 \right]\)}} \\
    \(\vec{\boldsymbol{K}}\) & \mtmain{Skeleton offsets.} \mtsub{It represents the offset of each point in the skeleton map between two time steps, \ie, \(\boldsymbol{K}_{t+1} - \boldsymbol{K}_{t}\)}. & \mtcell{Matrix \\ {\footnotesize\(\left[ 210 \times 3 \right]\)}} \\
    \midrule
    \(\boldsymbol{e}\) & Human boundaries. & Curve \\
    \(\boldsymbol{s}\) & Edges of flow map. & Curve \\
    \(\mathcal{P}_{\boldsymbol{e}}\) & Human boundaries within patch \(\mathcal{P}\). & Curve \\
    \(\mathcal{P}_{\boldsymbol{s}}\) & Edges of flow map within patch \(\mathcal{P}\). & Curve \\
    \midrule
    \(p\) & Any point within the human body in the flow maps. & Point \\
    \(\hat{q}\) & The closest point within skeleton offsets \(\vec{\boldsymbol{K}}\) with the highest visibility towards \(p\).  & Point \\
    \(i\) & Any point on the edges of flow map \( \boldsymbol{s} \). & Point \\
    \(\hat{j}\) & The nearest point within human boundaries \(\boldsymbol{e}\) towards \(i\). & Point \\
    \(\boldsymbol{c}_{\mathcal{P}_{\boldsymbol{e}}}\) & The centroid of curve \(\mathcal{P}_{\boldsymbol{e}}\). & Point \\
    \(\boldsymbol{c}_{\mathcal{P}_{\boldsymbol{s}}}\) & The centroid of curve \(\mathcal{P}_{\boldsymbol{s}}\). & Point \\
    \midrule
    \(\boldsymbol{v_s}\) & \mtmain{Subject motion.} \mtsub{It represents the overall motion trend of subjects.} & \mtcell{Vector \\ {\footnotesize\(\left[ \Delta x, \Delta y \right]\)}} \\ 
    \(\boldsymbol{u_p}\) & The estimated flow \(\boldsymbol{M}\) at point \(p\). & \mtcell{Vector \\ {\footnotesize\(\left[ \Delta x, \Delta y \right]\)}} \\
    \(\boldsymbol{k_{\hat{q}}}\) & The skeleton offset \(\vec{\boldsymbol{K}}\) at point \(\hat{q}\). & \mtcell{Vector \\ {\footnotesize\(\left[ \Delta x, \Delta y \right]\)}} \\
    \midrule
    \(\Phi\) & \mtmain{\methodName{}'s network component.} \mtsub{It estimates world flow \(\boldsymbol{M_w}\) between consecutive frames \(\boldsymbol{X}_{t}\) and \(\boldsymbol{X}_{t+1}\).} & \mtcell{Network \\ {\footnotesize[Params: \(\approx 3.4M\)]}} \\ 
    \(\Psi\) & \mtmain{\methodName{}'s network component.} \mtsub{It estimates subject motion \(\boldsymbol{v_s}\) based on world flow \(\boldsymbol{M_w}\).} & \mtcell{Network \\ {\footnotesize[Params: \(\approx 2.1M\)]}} \\ 
    \midrule
    \(\mathcal{F}\) & \mtmain{Our skeleton constraint.} \mtsub{It ensures that each body point's movement adheres to kinematic constraints.} & Function \\ 
    \(\mathcal{G}\) & \mtmain{Our boundary constraint.} \mtsub{It aligns human shapes onto our estimated flow maps.} & Function \\ 
    \(\mathcal{F_A}\) & \mtmain{Angular constraint.} \mtsub{Component of skeleton constraint.} & Function \\
    \(\mathcal{F_I}\) & \mtmain{Intensity constraint.} \mtsub{Component of skeleton constraint.} & Function \\
    \(\mathcal{C}\) & Chamfer distance between two curves. & Function \\
    \(\mathcal{D}\) & Euclidean distance between two points. & Function \\
    \midrule
    \(\omega_{1}\) & Learnable parameters in \(\Phi\). & Parameters \\
    \(\omega_{2}\) & Learnable parameters in \(\Psi\). & Parameters \\
    \midrule
    \(\vartheta_{a}\) & Threshold for angular constraint \(\mathcal{F}_{A}\). & Constant \\
    \(\vartheta_{i}^{l}\) \(\vartheta_{i}^{h}\) & Low and high boundary threshold for intensity constraint \(\mathcal{F}_{I}\). & Constant \\
    \bottomrule
    \end{tabular}
}
    \caption{\mcaption{Symbols used in the main text.} \msubcaption{Additionally, the description includes its relevant information.}}
    \label{tab:vars_matrix}
\end{table}




\end{document}